\documentclass{article} 
\usepackage{iclr_fmwild,times}

\usepackage{amsmath,amsfonts,bm}

\def\eqref#1{equation~\ref{#1}}

\def\1{\bm{1}}

\DeclareMathAlphabet{\mathsfit}{\encodingdefault}{\sfdefault}{m}{sl}
\SetMathAlphabet{\mathsfit}{bold}{\encodingdefault}{\sfdefault}{bx}{n}

\DeclareMathOperator*{\argmax}{arg\,max}

\usepackage{hyperref}
\usepackage{url}
\usepackage{booktabs}
\usepackage{multirow}
\usepackage[mode=buildnew]{standalone}
\usepackage{soul}
\usepackage{tikz}
\usepackage{amsmath,amssymb}
\usetikzlibrary{positioning,shapes,arrows,backgrounds,calc}

\usepackage{listings}
\usepackage{algorithm}
\usepackage{algpseudocode}

\usepackage[suppress]{color-edits}
\addauthor{matthew}{red}
\addauthor{andrew}{blue}
\addauthor{jamie}{green}
\addauthor{evan}{magenta}

\usepackage{xspace}
\newcommand{\eg}{\emph{e.g.,}\xspace}

\title{Faster, Cheaper, Better: Multi-Objective \\Hyperparameter Optimization for LLM and RAG Systems}

\author{Matthew Barker \\
    Trustwise AI \\
    \texttt{matthew@trustwise.ai} \\
    \And
    Andrew Bell \\
    New York University \\
    \texttt{alb9742@nyu.edu} \\
    \And
    Evan Thomas \\
    Trustwise AI \\
    \texttt{evan@trustwise.ai} \\
    \And
    James Carr \\
    Trustwise AI \\
    \texttt{jamie@trustwise.ai} \\
    \And
    Thomas Andrews \\
    Trustwise AI \\
    \texttt{thomas@trustwise.ai} \\
    \And
    Umang Bhatt \\
    New York University \\
    The Alan Turing Institute\\
    \texttt{umangbhatt@nyu.edu} \\
}

\iclrfinalcopy 
\begin{document}

\maketitle

\begin{abstract}
While Retrieval Augmented Generation (RAG) has emerged as a popular technique for improving Large Language Model (LLM) systems, it introduces a large number of choices, parameters and hyperparameters that must be made or tuned. This includes the LLM, embedding, and ranker models themselves, as well as hyperparameters governing individual RAG components. Yet,  collectively optimizing the entire configuration in a RAG or LLM system remains under-explored---especially in multi-objective settings---due to intractably large solution spaces, noisy objective evaluations, and the high cost of evaluations.

In this work, we introduce the first approach for multi-objective parameter optimization of cost, latency, safety and alignment over entire LLM and RAG systems. We find that Bayesian optimization methods significantly outperform baseline approaches, obtaining a superior Pareto front on two new RAG benchmark tasks. We conclude our work with important considerations for practitioners who are designing multi-objective RAG systems, highlighting nuances such as how optimal configurations may not generalize across tasks and objectives.
\end{abstract}

\section{Introduction}
Retrieval Augmented Generation (RAG) has emerged as a popular technique for improving the performance of Large Language Models (LLMs) on question-answering tasks over specific datasets. A benefit of using RAG pipelines is that they can often achieve high performance on specific tasks \emph{without} the need for extensive alignment and fine-tuning~\citep{gupta2024rag}, a costly and time-consuming process. However, the end-to-end pipeline of a RAG system is dependent on many parameters that span different components (or modules) of the system, such as the choice of LLM, the embedding model used in retrieval, the number of chunks retrieved and hyperparameters governing a reranking model. Examples of choices, parameters, and hyperparameters that are often made or tuned when implementing a RAG pipeline are listed in Table~\ref{tab:system_params}. Importantly, the performance of a RAG pipeline is dependent on these choices~\citep{fu2024autorag}, many of which can be difficult to tune manually. While those building RAG pipelines might avoid fine-tuning costs, they often spend time and resources on hyperparameter optimization (HO).

Despite this, there is little research exploring methods for collectively optimizing all the hyperparameters in a given LLM and RAG pipeline~\citep{fu2024autorag}. Further, to the best of our knowledge, there is no work that addresses this challenge in \emph{multi-objective settings}, where the RAG pipeline must achieve high performance across a range of objectives,
like 
minimizing a system's inference time while maximizing its helpfulness. In this work, we aim to fill this gap by introducing an approach for collectively optimizing the hyperparameters of a RAG system in a multi-objective setting.

The authors of this paper are from a mixture of both academia and industry, and this work was motivated by real-world challenges faced by industry practitioners. Use of RAG pipelines within industry often requires balancing multiple requirements which are in competition with one another. For example, at one financial services firm developing an in-house Q\&A chatbot to support internal workflows, practitioners aimed to both maximize accuracy and minimize the time taken to generate a response. However, there is a tension in these two objectives: a RAG system utilizing larger models may yield more accurate responses, but consequently requires longer computation time. As another example, a large bank developing an external insurance policy Q\&A chatbot was primarily concerned with the alignment of generated responses to policies and regulations. Objectives that we have observed practitioners frequently consider when building RAG pipelines include: cost, response latency, safety (hallucination risk), and alignment (response helpfulness).

Multi-objective HO over a RAG pipeline is particularly challenging for several reasons. First, RAG pipelines naturally have a high number of parameters, leading to a large solution space. We identify at least 15 example choices, parameters, and hyperparameters in Table~\ref{tab:system_params}. Even if one has just a handful of possible values for each choice, the parameter space becomes intractably large for simple algorithms like grid search~\citep{bergstra2012random}. Second, evaluating a RAG pipeline during the HO process is costly, with respect to both compute resources and time: it requires running the RAG system over multiple queries (where each iteration is bounded by the per-token inference time of the LLM), and then evaluating each output. Third, the objective evaluations can be noisy. The true characteristic of a RAG system cannot be computed directly, and requires sampling the evaluations from many queries. Relatedly, since in most cases LLMs are non-deterministic, combinations of hyperparameters need to be tested over multiple seeds.

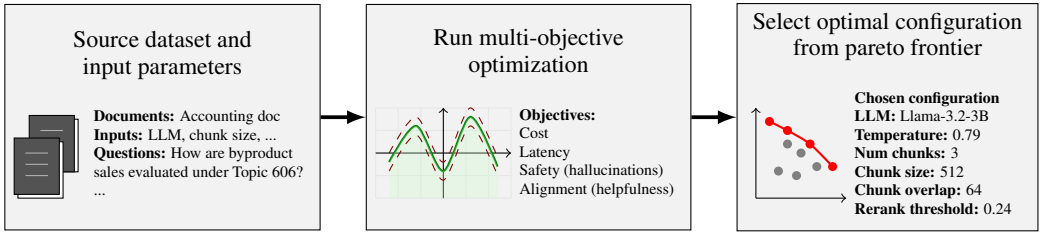
\begin{figure}[t]
    \centering  \begin{tikzpicture}[
    box/.style={
        draw,
        rectangle,
        minimum width=4cm,
        minimum height=3cm,
        align=center,
        fill=gray!10,
        font=\small,
    },
    arrow/.style={
        ->,
        line width=0.5mm,
        >=latex
    }
]

\node[box] (data) {Source dataset and\\input parameters\\ \\
    \begin{tikzpicture}[scale=0.5]
        \draw[fill=white] (0.1,0.1) -- (1.1,0.1) -- (1.1,1.4) -- (1.3,1.4) -- (1.3,0) -- (0.1,0) -- cycle;
        \draw[fill=darkgray!90] (0,0) rectangle (1.2,1.5);
        \draw[black!30] (0.3,1.1) -- (0.9,1.1);
        \draw[black!30] (0.3,0.8) -- (0.9,0.8);
        \draw[black!30] (0.3,0.5) -- (0.9,0.5);
        \draw[fill=white] (-0.3,-0.5) -- (0.7,-0.5) -- (0.7,0.8) -- (0.9,0.8) -- (0.9,-0.7) -- (-0.3,-0.7) -- cycle;
        \draw[fill=darkgray!90] (-0.4,-0.6) rectangle (0.8,0.9);
        \draw[black!30] (-0.1,0.5) -- (0.5,0.5);
        \draw[black!30] (-0.1,0.2) -- (0.5,0.2);
        \draw[black!30] (-0.1,-0.1) -- (0.5,-0.1);

        \node[align=left, minimum width=3cm, minimum height=1cm, font=\tiny] at (4.5,0.5) {\textbf{Documents:} Accounting doc\\ \textbf{Inputs:} LLM, chunk size, ... \\ \textbf{Questions:} How are byproduct\\sales evaluated under Topic 606?\\...};
    \end{tikzpicture}
};

\node[box, right=0.6cm of data] (opt) {Run multi-objective\\optimization\\ \\
    \begin{tikzpicture}[scale=0.6]
        \draw[gray!20] (-1.5,-1) grid[step=0.5] (1.5,1);
        \draw[->] (-1.5,0) -- (1.5,0);
        \draw[->] (0,-1) -- (0,1);
        \draw[green!50!black, thick] plot [smooth] coordinates {(-1.2,-0.2) (-0.6,0.6) (0,-0.4) (0.6,0.8) (1.2,-0.3)};
        \fill[green!20, opacity=0.3] plot [smooth] coordinates {(-1.2,-0.2) (-0.6,0.6) (0,-0.4) (0.6,0.8) (1.2,-0.3)} -- (1.2,-1) -- (-1.2,-1) -- cycle;
        \draw[red!50!black, dashed] plot [smooth] coordinates {(-1.2,0) (-0.6,0.8) (0,-0.2) (0.6,1) (1.2,-0.1)};
        \draw[red!50!black, dashed] plot [smooth] coordinates {(-1.2,-0.4) (-0.6,0.4) (0,-0.6) (0.6,0.6) (1.2,-0.5)};

        \node[align=left, minimum width=2cm, minimum height=1cm, font=\tiny] at (1.5,0) {\textbf{Objectives:}\\Cost\\Latency\\Safety (hallucinations)\\Alignment (helpfulness)};
    \end{tikzpicture}
};

\node[box, right=0.6cm of opt] (pareto) {Select optimal configuration\\from pareto frontier\\ \\
    \begin{tikzpicture}[scale=0.6]
        \draw[->] (-1,-1) -- (1,-1);
        \draw[->] (-1,-1) -- (-1,1);
        \foreach \x/\y in {-0.5/-0.4, -0.1/-0.5, 0/0, -0.3/0.2, 0.35/-0.3}
            \fill[gray] (\x,\y) circle (0.1);
        \draw[red, thick] plot [smooth] coordinates {(-0.7,0.7) (-0.3,0.5) (0.2,0.2) (0.7,-0.3)};
        \foreach \x/\y in {-0.7/0.7, -0.3/0.5, 0.2/0.2, 0.7/-0.3}
            \fill[red] (\x,\y) circle (0.1);

        \node[align=left, minimum width=1.5cm, minimum height=1cm, font=\tiny] at (1,0) {\textbf{Chosen configuration}\\ \textbf{LLM:} Llama-3.2-3B\\ \textbf{Temperature:} 0.79 \\ \textbf{Num chunks:} 3\\ \textbf{Chunk size:} 512\\ \textbf{Chunk overlap:} 64 \\ \textbf{Rerank threshold:} 0.24};

    \end{tikzpicture}
};

\draw[arrow] (data) -- (opt);
\draw[arrow] (opt) -- (pareto);

\end{tikzpicture}
    \caption{A high-level overview of our approach. First, we source the datasets that we will use to optimize our RAG pipeline, define the choices, parameters and hyperparameters that will be optimized over (see Table~\ref{tab:system_params}), and select the objectives for optimization (\eg cost, latency, safety, and alignment). 
    Second, we introduce a train-test paradigm for evaluating the performance of RAG pipelines, and use Bayesian optimization (BO) to find the optimal parameter configurations. We find that using BO with the \texttt{qLogNEHVI}~\citep{daulton2021parallel, ament2023unexpected} acquisition function is well-suited for this problem, since it is adapted for noisy objective evaluations and makes use of a single composite objective called \emph{hypervolume improvement} that allows for an arbitrary number of objectives. Third, we explore the Pareto frontier of parameter configurations, finding the best solutions over different objectives.}
    \label{fig:enter-label}
\end{figure}

\paragraph{Contributions.} In this work, we make four main contributions:

\begin{enumerate}
    \item[(1)] We introduce an approach for multi-objective optimization over a unique set of hyperparameters of a RAG pipeline, including choices for the LLM and embedding models themselves. Our approach implements a single composite objective value called the hypervolume indicator~\citep{guerreiro2021hypervolume}, and uses Bayesian optimization with an acquisition function that allows for an arbitrary number of noisy objective functions (\texttt{qLogNEHVI}) ~\citep{daulton2021parallel, ament2023unexpected} to find the best RAG pipeline configuration.
    
    \item[(2)] We empirically show the effectiveness of our approach to identify optimal RAG pipeline configurations across two tasks (one related to financial services Q\&A, and another related to medical Q\&A) using a train-test paradigm, as compared to random parameter choices and other baseline optimization approaches.

    \item[(3)] We publicly release two novel benchmarks for evaluating RAG systems called \texttt{FinancialQA} and \texttt{MedicalQA}.\footnote{\url{https://huggingface.co/datasets/Trustwise/optimization-benchmark-dataset}} Importantly, these benchmarks more closely mimic industry RAG use-cases than currently available benchmarks, since the context must be retrieved at runtime from an available document, rather than being given in the dataset.
    
    \item[(4)] We frame our discussion (Section~\ref{sec:discussion}) as guidance to practitioners who seek to improve their own RAG systems. We highlight two important considerations: the first is what we call ``task dependence'', meaning that an optimal configuration for a RAG pipeline on a task in a specific setting may not generalize to another setting. The second is ``objective dependence'', where objective evaluations follow different trends (or have no trend) across different configurations. Task and objective dependence can also compound, highlighting the challenge of collectively optimizing the parameters of a RAG system.
\end{enumerate}

\section{Related work}
\label{sec:related}
There has been a mixture of work separately addressing multi-objective optimization and hyperparameter optimization in LLM and RAG systems, which we summarize here. We provide further comparison with fine-tuning and model-merging approaches in Appendix ~\ref{appendix:related}.

\paragraph{Hyperparameter optimization (HO).} As many pieces of LLM and RAG pipelines have hyperparameters that must be tuned before deployment, there is a large body of work testing the efficacy of using HO to tune these systems. 

\citet{wang2023cost} propose a cost-based pruning strategy to hyperparameter tune LLM systems under budget constraints. They focus on hyperparameters like the type of model (\eg text-davinci-003, gpt-3.5-turbo, or gpt-4), the maximum number of tokens that can be generated in a response, the model temperature\footnote{A parameter for which low or high values sharpen or soften the probability distribution of a token being outputted by an LLM, respectively.}, and the model top-$p$.\footnote{A parameter that restricts the domain of tokens that can be outputted by an LLM to those whose cumulative probability is greater than $p$.} They use a search method called BlendSearch~\citep{wang2021economic}, which combines Bayesian optimization and local search~\citep{wu2021frugal}, to find the optimal combinations of parameters, and measure performance on the tasks APPS~\citep{hendrycks2021measuring}, XSum~\citep{narayan2018don}, MATH~\citep{hendrycks2021measuring}, and HumanEval~\cite{chen2021evaluating}. In comparison with our work, \citet{wang2023cost} do not consider a RAG system.

Most related to our work, \citet{kim2024autorag} proposed AutoRAG, an open-source framework designed for RAG experimentation and hyperparameter optimization. They use a a greedy algorithm for selecting the hyperparameters governing RAG modules like query expansion, retrieval, passage augmentation, passage re-ranking, prompt making, and generating (the LLM). In concurrent work, \citet{fu2024autorag} proposed AutoRAG-HP, which frames hyperparameter selection as an online multi-armed bandit (MAB) problem. To carry out HO, they introduce a novel two-level Hierarchical MAB (Hier-MAB) method, where a high-level MAB guides the optimization of modules, and several low-level MABs search for optimal settings within each module. Significantly, our work is distinct from both \citet{kim2024autorag} and \citet{fu2024autorag} in that they do not consider multi-objective settings.

\paragraph{Multi-objective alignment.} Several researchers have proposed methods for incorporating multiple objectives directly into the LLM fine-tuning and alignment processes. ~\citet{li2020deep} developed an approach for multi-objective alignment from human feedback using scalar linearization. ~\cite{mukherjee2024multi} expanded on that approach by developing an algorithm that finds a diverse set of Pareto-optimal solutions that maximize the hypervolume, given a set of objectives.~\cite{zhou2024beyond} proposed a reward-function free extension called Multi-Objective Direct Preference Optimization (MODPO). The latter showed that MODPO can effectively find a Pareto-optimal frontier of fine-tuned models, trading off objectives like ``helpfulness'' and ``harmlessness''. While these works have demonstrated success in multi-objective LLM alignment, we focus on RAG pipelines and avoid aligning and fine-tuning models altogether. 
\begin{table}[t]
\caption{Common choices, parameters, and hyperparameters that are often made/tuned when implementing RAG pipelines. \textbf{Bold} indicates a parameter that was optimized over in our experiments.}
\label{tab:system_params}
\begin{tabular}{p{2cm}p{3.4cm}p{7.1cm}}
\toprule
\textbf{Domain} & \textbf{Parameters} & \textbf{Notes} \\
\midrule
\multirow{2}{*}{System-level} & \textbf{LLM model} & \eg gpt-4, llama-3.1-8b, llama-3.1-70b \\
 & \textbf{Embedding model} & For RAG pipelines, \eg text-embedding-ada-002 \\
\midrule
\multirow{4}{*}{LLM controls} & System prompt &  \\
 & \textbf{Temperature} &  \\
 & Top-$k$, top-$p$ &  \\
 & Max length of output &  \\
\midrule
\multirow{9}{*}{Fine-tuning} & Preference-tuning approach & \eg RLHF, DPO, KTO; these methods may also introduce hyperparameters that must be tuned, \eg DPO uses $\beta$, KTO uses $\beta, \lambda_U, \lambda_D$ \\
 & Parameter-Efficient Fine-Tuning (PEFT) & \eg LoRA, adapter modules~\citep{houlsby2019parameter}; these methods may also introduce hyperparameters, \eg LoRA uses rank and scaling $\alpha$ \\
 & Dropout rate &  \\
 & Learning rate &  \\
 & Training epochs &  \\
\midrule
\multirow{7}{*}{RAG controls}
& Modules & Approaches for query expansion, retrieval, passage augmentation and re-ranking, and prompt making
\\
& \textbf{Chunk size} &  \\
& \textbf{Number of chunks} & \\
& \textbf{Chunk overlap} & \\
& \textbf{Re-rank threshold} \\
& Retrieval size (top-$k$) & \\
\bottomrule
\end{tabular}
\end{table}

\section{Preliminaries and problem formulation}
\label{sec:prelims}

\paragraph{Multi-Objective Optimization (MOO).}
The goal of MOO is to find a solution $\mathbf{x} \in \mathcal{X}$ that maximizes (or minimizes) a set of objective functions, where $\mathbf{x} = [x_1, x_2, \dots x_l]$ corresponds to a series of input values, and $\mathcal{X}$ is said to be the solution space. We then define each objective as $f : \mathcal{X} \rightarrow \mathbb{R}$ and use $\mathbf{f} : \mathcal{X} \rightarrow\mathbb{R}^k$ to represent $k$ objective functions. Using these definitions, we aim to solve the following optimization problem:

\begin{equation}
\max_{\mathbf{x} \in \mathcal{X}} ~\mathbf{f(x)} := \max_{\mathbf{x} \in \mathcal{X}} \left[ f_1(\mathbf{x}),f_2(\mathbf{x}), \dots f_k(\mathbf{x})\right]
\label{eq:moo}
\end{equation}

Rather than identifying a single solution, MOO algorithms identify a set of non-dominated solutions~\citep{deb2016multi}. We use $\mathbf{f(x^*)} \succ \mathbf{f(x)}$ to signify that $\mathbf{f(x^*)}$ dominates $\mathbf{f(x)}$:
\begin{equation}
\forall i \in \{1, \dots, k\}, \quad f_i(\mathbf{x}) \leq f_i(\mathbf{x^*}),  \quad \text{and}\quad  \exists j \text{ s.t. } f_j(\mathbf{x}) < f_j(\mathbf{x^*})
\end{equation}
Hence, and following \cite{daulton2021parallel}, we define the \emph{Pareto set} as $\mathcal{P}^* = \{ \mathbf{f(x^*)} \mid \mathbf{x^*}\in \mathcal{X}, \nexists \; \mathbf{x} \in \mathcal{X} \text{ s.t. } \mathbf{f(x)} \succ \mathbf{f(x^*)} \}$, and the corresponding \emph{Pareto optimal solutions} as $\mathcal{X^*} = \{ \mathbf{x^*} \mid \mathbf{f(x^*)} \in \mathcal{P}^*\}$. In practice, the Pareto optimal set often consists of an infinite set of points. Given a set of observed solutions from $\mathcal{X}$, 
we aim to identify an approximate Pareto optimal set, $\mathcal{\hat{P}} \subset \mathbb{R}^k$, and its associated Pareto optimal solutions, $\mathcal{\hat{X}}$. We then use the hypervolume (HV) indicator, $\mathcal{HV}$, to evaluate the quality of $\mathcal{\hat{P}}$ given a reference point $\mathbf{r} \in \mathbb{R}^k$. Our optimization problem then becomes:
\begin{equation}
    \label{eq:objective}
    \argmax_{\mathcal{\hat{P}}} \mathcal{HV}(\mathcal{\hat{P}}|\mathbf{r})
\end{equation}
In this work, we seek to find the Pareto-optimal combinations of parameters of a RAG system (those indicated in Table~\ref{tab:system_params}). We represent a \emph{configuration} of a RAG system as a solution vector $\mathbf{x} \in \mathcal{X}$, where each value corresponds to a system parameter. We aim to find the solution set containing different RAG pipeline configurations that maximizes $\mathcal{HV}$ for the objectives in Section \ref{sec:objectives}.

\paragraph{Bayesian Optimization (BO).} Objective function evaluations herein are obtained using an entire RAG pipeline, meaning there is no single analytic expression or gradient available for solving Equation (\ref{eq:objective}). BO is a derivative-free optimization method that works by constructing probabilistic surrogate models to capture the uncertainty of objective functions. The core methodology employs Gaussian process regression to model the unknown objective landscape, where each function $f_i$ is represented as a stochastic process with a posterior distribution $p(f_i \mid \mathcal{D})$ conditioned on observed data \citep{williams2006gaussian}. Importantly, BO makes use of an acquisition function to balance exploration of unknown regions and exploitation of promising solutions. BO has been known to perform well compared to other optimization methods when objective functions are expensive to evaluate, as in our setting~\citep{gramacy2020surrogates, diessner2022investigating,guerreiro2021hypervolume}.

\section{Methodology}
\label{sec:methodology}

Our approach works by defining a solution space (over the configurations of a RAG pipeline), objectives, and a train-test paradigm, then using BO to find the optimal configuration. BO is well-suited to exploit patterns in objective evaluations, for example the tendency for latency to increase with chunk size. We allow that the solution space has a mixture of continuous variables (\eg temperature of LLM) and categorical variables (\eg choice of LLM). In addition, we allow for constraints on the inputs, such as asserting that the chunk overlap must be less than the chunk size.

We make use of two state-of-the-art algorithms that implement and extend BO. The first, \texttt{qLogEHVI}, takes advantage of recent advances in programming models and hardware acceleration to parallelize multi-objective BO using the LogEI variant~\citep{ament2023unexpected} of \emph{expected hypervolume improvement} to guide the acquisition of new candidate solutions~\citep{daulton2020differentiable}. The second is \texttt{qLogNEHVI}, which extends \texttt{qLogNEHVI} by using a novel acquisition function that was theoretically motivated and empirically demonstrated to outperform benchmark methods in settings with noisy objective evaluations~\citep{daulton2021parallel}. The noisy variant is particularly useful since the probabilistic nature of LLMs can cause noisy objective function evaluations.
BO is also well-suited to exploit patterns in objective evaluations, for example the tendency for latency to increase with chunk size. Following \cite{daulton2021parallel}, we initialize both BO algorithms with $N_\text{init}$ points from a scrambled Sobol sequence.

We outline our proposed methodology in Algorithm \ref{alg:1}, and provide implementation details in Section~\ref{sec:experiments}. We also provide a method of approximating each objective function in Appendix \ref{sec:approx_objective}.

\begin{algorithm}
\caption{Train-test multi-objective optimization of RAG or LLM system}\label{alg:1}
\begin{algorithmic}[1]
\Require set of documents $\mathcal{D}$, solution space $\mathcal{X}$, reference point $\mathbf{r}$, number of iterations $N$, number of iterations for BO initialization $N_{\text{init}}$ 
\State $Q, Q_{\text{test}} \gets \texttt{Generate}(\mathcal{D})$ \Comment{Generate train-test queries from documents}
\State $H, H_{\text{test}}\gets [~], [~]$ \Comment{Start with empty train and test history}

\For{$n = 1...N$} \Comment{Parameter optimization}
\If{$n\leq N_{\text{init}}$}
    \State $\mathbf{x} \gets \texttt{Sobol}(H, \mathcal{X})$ \Comment{Sobol sampling for $N_{\text{init}}$ iterations}
\Else
    \State $\mathbf{x} \gets \texttt{qLogNEHVI}(H, \mathcal{X})$ \Comment{\texttt{qLogNEHVI} acquisition function}
\EndIf
\State $\mathbf{f} \gets  \texttt{ObjectiveEvaluations}(\mathbf{x}, Q)$ \Comment{Evaluate solution on $Q$}
\State $H.\texttt{append} (\{\mathbf{x}, \mathbf{f}\})$ \Comment{Save to history}

\State $\mathcal{\hat{X}}, \mathcal{\hat{P}} \gets \texttt{ParetoOptimalSet} (H)$ \Comment{Find Pareto-optimal set}
\State $\text{HV} \gets \mathcal{HV} (\mathcal{\hat{P}} | \mathbf{r})$ \Comment{Find hypervolume w.r.t reference point}

\State $\mathbf{f}_{\text{test}} \gets \texttt{ObjectiveEvaluations}(\mathbf{x}, Q_{\text{test}})$ \Comment{Evaluate solution on $Q_{\text{test}}$}
\State $H_{\text{test}}.\texttt{append}(\{ \mathbf{x}, \mathbf{f}_{\text{test}}\})$ \Comment{Save to history}
\State $\mathcal{\hat{X}}_\text{test}, \mathcal{\hat{P}}_\text{test} \gets \texttt{ParetoOptimalSet} (H_{\text{test}})$ \Comment{Find test Pareto-optimal set}
\State $\text{HV}_\text{test} \gets \mathcal{HV} (\mathcal{\hat{P}}_{\text{test}} | \mathbf{r})$ \Comment{Find test hypervolume w.r.t reference point}
\EndFor
\State $X_{\text{opt}} \gets \texttt{SelectOptimalConfig}(\hat{\mathcal{X}}_{\text{test}},\mathcal{\hat{P}}_{\text{test}})$ \Comment{Select optimal configuration(s)}
\end{algorithmic}
\end{algorithm}

\subsection{Generating Train-Test Queries}
\label{sec:synthetic_data}
A frequent challenge encountered by the authors in industry is a lack of existing queries for RAG Q\&A tasks.
To this end, we use LLMs to help generate queries from the data, which may be PDF documents or large documents of text. LLMs are a reasonable choice for this approach because they have been shown to be effective at generating synthetic data \citep{long2024llms}.
For our \texttt{FinancialQA} dataset described in Section \ref{sec:datasets}, we generate train and test queries using an LLM (GPT-4o). We publicly release the synthetic questions as part of our datasets, and provide the prompt used to generate the questions in Appendix \ref{sec:prompts}.

\subsection{Objectives}
\label{sec:objectives}

Motivated by experiences in industry, we consider four objectives that practitioners commonly consider important: safety, alignment, cost and latency.

\paragraph{Safety.} In this work, we use the term ``safety'' to refer to \emph{hallucination risks}, or the risk that a RAG pipeline will return false information to the user. Hallucinations can cause significant downstream harm, particularly in high-stakes domains such as healthcare.
In our experiments, we evaluate safety using the \emph{faithfulness} metric defined in the Trustwise API.\footnote{\label{docnote} Trustwise documentation available at: \url{https://trustwise.ai/docs}} Like previous work \citep{min2023factscore, es2023ragas}, faithfulness detects hallucinations by evaluating whether or not the response from a RAG system is supported by the context. The response is split into individual, ``atomic'' claims that are verified with respect to the context. Scores of these verifications are then aggregated into a single faithfulness score between 0 and 100 for each response, where 100 represents a completely ``safe'' response with no hallucinations.\footnote{For those aiming to replicate our approach, there are open-source alternatives for evaluating safety that could be used instead, such as LlamaGuard~\citep{inan2023llama}.}

\paragraph{Alignment.}
While hallucination risks are an immediate concern, the alignment of a response is often just as important in enterprise use-cases. To evaluate alignment, we follow the definition of \emph{helpfulness} popularized by Anthropic~\citep{bai2022training}. We measure alignment using the \emph{helpfulness} metric as implemented in the Trustwise API which judges how useful, detailed and unambiguous a response is. This metric assigns a score between 0 and 100 for each response, where a higher score indicates a more helpful response.

\paragraph{Cost.} To calculate the cost of an evaluation, we consider all the components of a RAG pipeline, including the query embedding cost, reranker embedding cost, LLM input token cost and LLM output token cost. Importantly, the cost of a RAG pipeline is a function of its configuration:
\begin{align}
\begin{split}
    \text{cost} = ~&\text{number of query tokens} \times \text{cost per embedding token} \\
    &+\text{number of context tokens} \times \text{cost per reranker token} \\
    &+ \text{number of prompt input tokens} \times \text{cost per LLM input token} \\
    &+ \text{number of output tokens} \times \text{cost per LLM output token}
\end{split}
\end{align}

The embedding cost per query token, reranker token cost, LLM input token cost, and LLM output token cost are based on the specific choices of those models, as well as the hardware being used to run the RAG pipeline. In enterprise use-cases, these costs may also include overhead and maintenance.

\paragraph{Latency.} We define latency as the time it takes for a complete end-to-end run of the RAG pipeline, from the moment an initial query is sent to the system to the moment a full response is returned to the user. As with cost, we can calculate the latency of a system as a function of its configuration:
\begin{equation}
    \text{latency} = \text{embedding latency} + \text{reranker latency} + \text{LLM response latency} + \text{evaluation latency}
\end{equation}
We note that response evaluations can take as long as, or longer than, response generation and thus the end-to-latency is vital to consider in enterprise settings where evaluations are a requirement.
\section{Experiments}
\label{sec:experiments}

We tested our optimization approaches on two RAG tasks from different industries.  We use a standard RAG setup with retrieval using vector embeddings, a reranker to filter out unnecessary context chunks, and an LLM prompted to generate a response using the context. The exact configuration solution space we use is given in Appendix~\ref{appendix:experiments}.

We optimize for the four objectives of cost, latency, safety and alignment. We run BO using the train question set, and then report results on a held out test set. We use \emph{Ax} \citep{bakshy2018ae} and \emph{BoTorch} \citep{balandat2020botorch} to run and manage experiments. For all algorithms we use 50 total iterations, and for BO methods, the first 20 iterations are chosen using Sobol sampling. In MOO, a reference point is used to calculate the HV improvement, and represents the minimum acceptable solution. Based on our industry experience, we use a reference point with cost of \$2000 per million queries, latency of 20s per query, safety of 50 and alignment of 50. This choice of reference point prevents degenerate solutions (\eg a degenerate RAG system which does not retrieve any chunks) from contributing to the HV improvement.

\textit{Throughout this work, we report the cost in USD per million queries (\$ / million queries) and latency in seconds (s). Safety and alignment scores are dimensionless and range from 0 to 100. Cost and latency are objectives to minimize, while safety and alignment are objectives to be maximized.}

\begin{figure}[t]
    \centering
    \includegraphics[width=\linewidth]{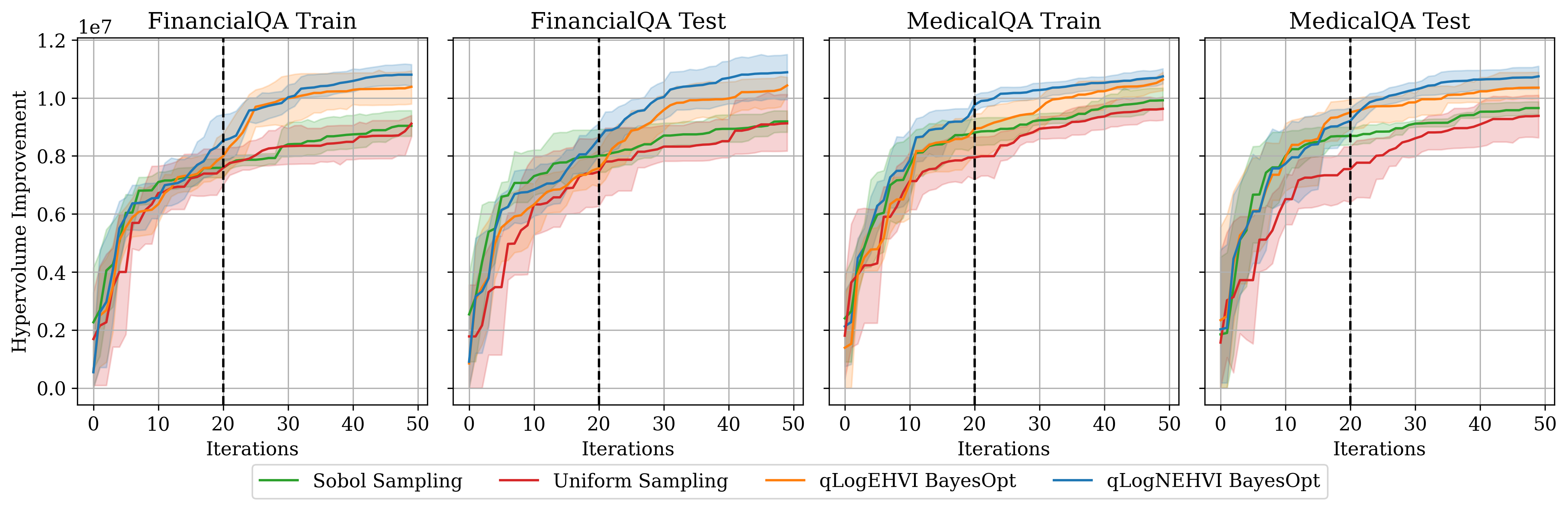}
    \caption{HV improvement on train and test splits for both datasets. Our proposed acquisition function for BO (\texttt{qLogNEHVI}) outperforms its noiseless variant (\texttt{qLogEHVI}) and both BO algorithms perform significantly better than the baselines. There is a noticeable increase in HV after iteration 20 (dotted line), indicating the end of Sobol sampling initializations for the BO algorithms, and the start of acquisition function-guided selections.}
    \label{fig:hvi}
\end{figure}

\subsection{Datasets}
\label{sec:datasets}
Existing datasets for Q\&A tasks generally exist in the form of (question, context, answer) triplets. However, in industry RAG use-cases, the context is often retrieved at runtime from a set of documents. As such, we want to include the retrieval of the context as part of of our evaluation. To this end, we adapt two known tasks to fit the needs of our experimental setup\footnote{We publicly release the train and test questions for these documents at \url{https://huggingface.co/datasets/Trustwise/optimization-benchmark-dataset}}:
\paragraph{FinancialQA.}
This task uses a publicly available document covering revenue recognition from a leading global accounting firm.\footnote{Financial document available at: \url{https://kpmg.com/kpmg-us/content/dam/kpmg/frv/pdf/2024/handbook-revenue-recognition-1224.pdf}} The document includes more than 1000 pages of text, representing a significant Q\&A context retrieval challenge. Since the document does not come with a set of questions, we synthetically generate 50 questions by prompting GPT-4o, using the method described in Section~\ref{sec:synthetic_data}. We then randomly split this set into 30 train and 20 test questions.
\paragraph{MedicalQA.}
We create our medical dataset using the existing FACTS benchmark \citep{jacovi2025facts} which includes (question, context, answer) triplets.\footnote{FACTS dataset available at: \url{https://www.kaggle.com/facts-leaderboard}} We take the medical Q\&A subset, and the authors manually filter out unrepresentative questions. We then combine all the individual context chunks from each question in the medical Q\&A split into one large document. The final dataset includes one large medical document, and independent sets of 43 train and 43 test questions.

\subsection{Baselines}
For our tasks, we lack access to a ``ground truth'' set of Pareto optimal configurations because we cannot evaluate the objective functions directly, and grid search is computationally unfeasible. Hence we use three baseline approaches that are applicable in MOO settings. The first is \emph{uniform sampling}, which generates configurations independently, where each configuration from the solution space is equally probable. 
The second is \emph{Sobol sampling.} Like Latin hypercube sampling~\citep{loh1996latin},  this method generates configurations that guarantee good high-dimensional uniformity. It is commonly used to initialize optimization algorithms, including BO, and represents a sensible alternative to grid-search. Finally, we use \texttt{qLogEHVI} BO, the noiseless variant of our chosen acquisition function \citep{daulton2020differentiable}.

\section{Results}
\label{sec:results}

\begin{figure}[t]
    \centering
    \includegraphics[width=\linewidth]{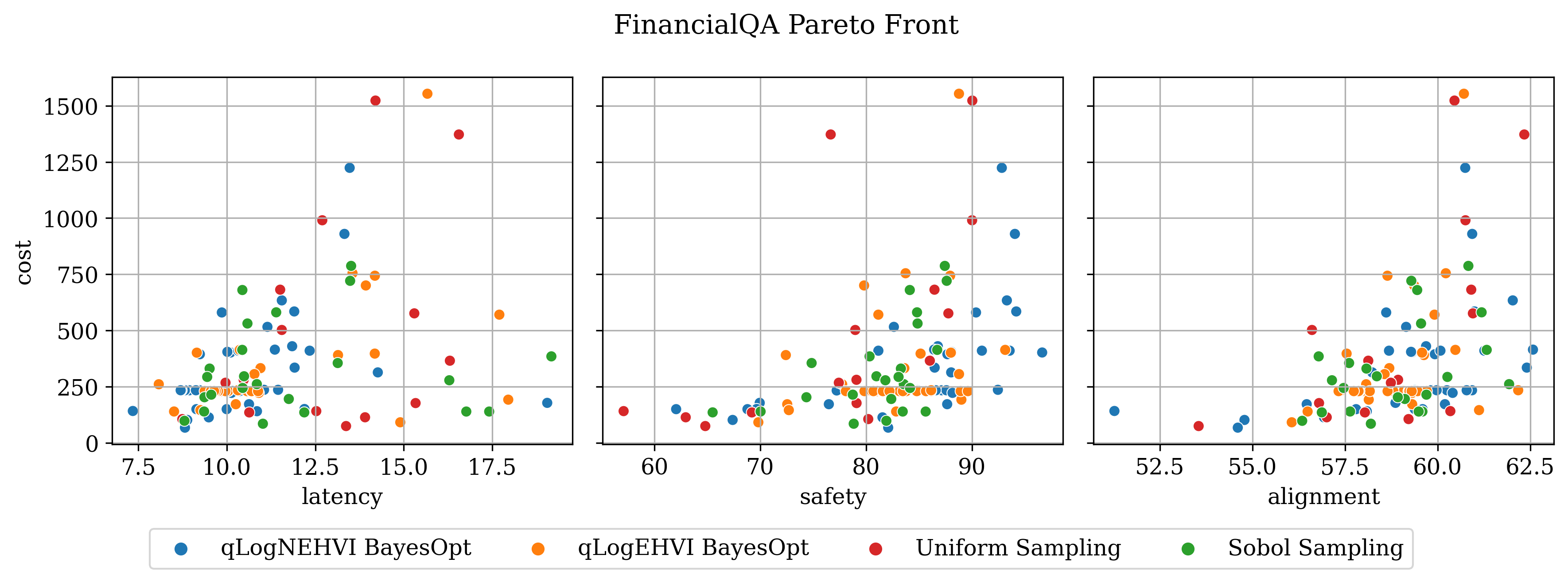}
    \includegraphics[width=\linewidth]{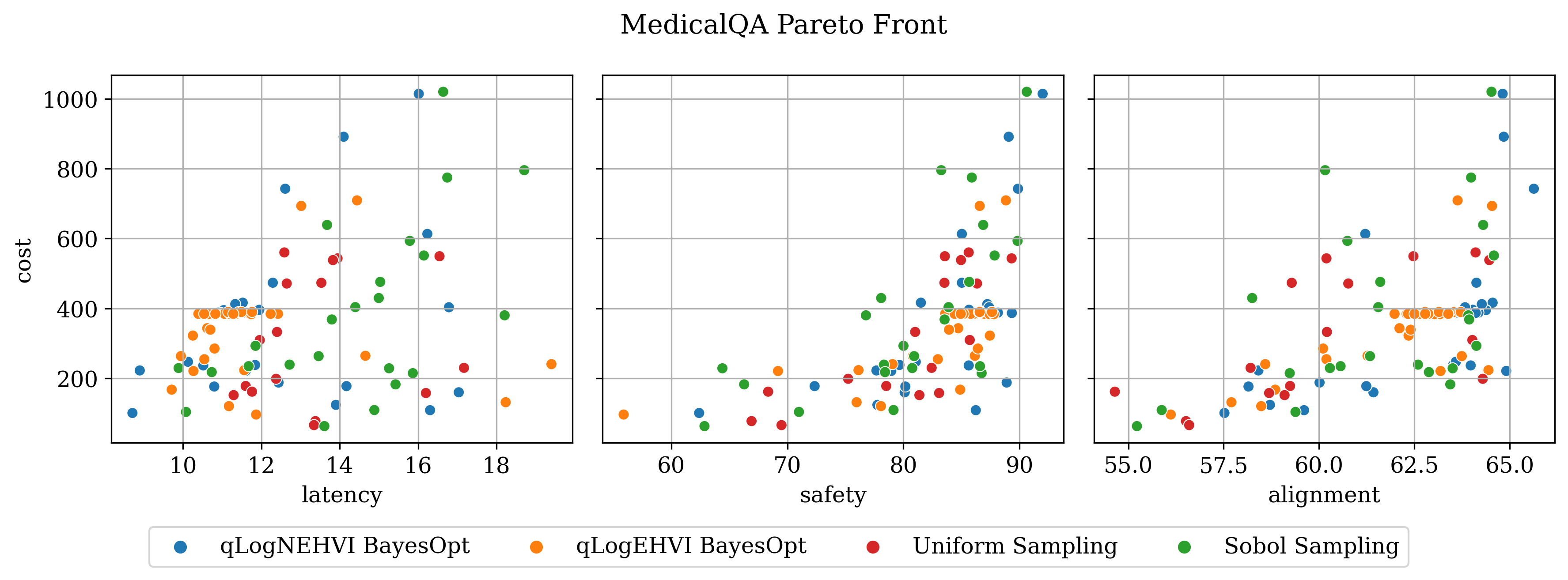}
    \caption{2D projections of the 4D Pareto frontier for each algorithm for a fixed random seed on both datasets. We see our proposed algorithm (\texttt{qLogNEHVI} BayesOpt) obtains a superior Pareto front, with solutions concentrated towards high safety, high alignment, low cost, and low latency.}
    \label{fig:pareto-front}
\end{figure}
\begin{figure}[t]
\centering
   \includegraphics[width=0.4\linewidth]{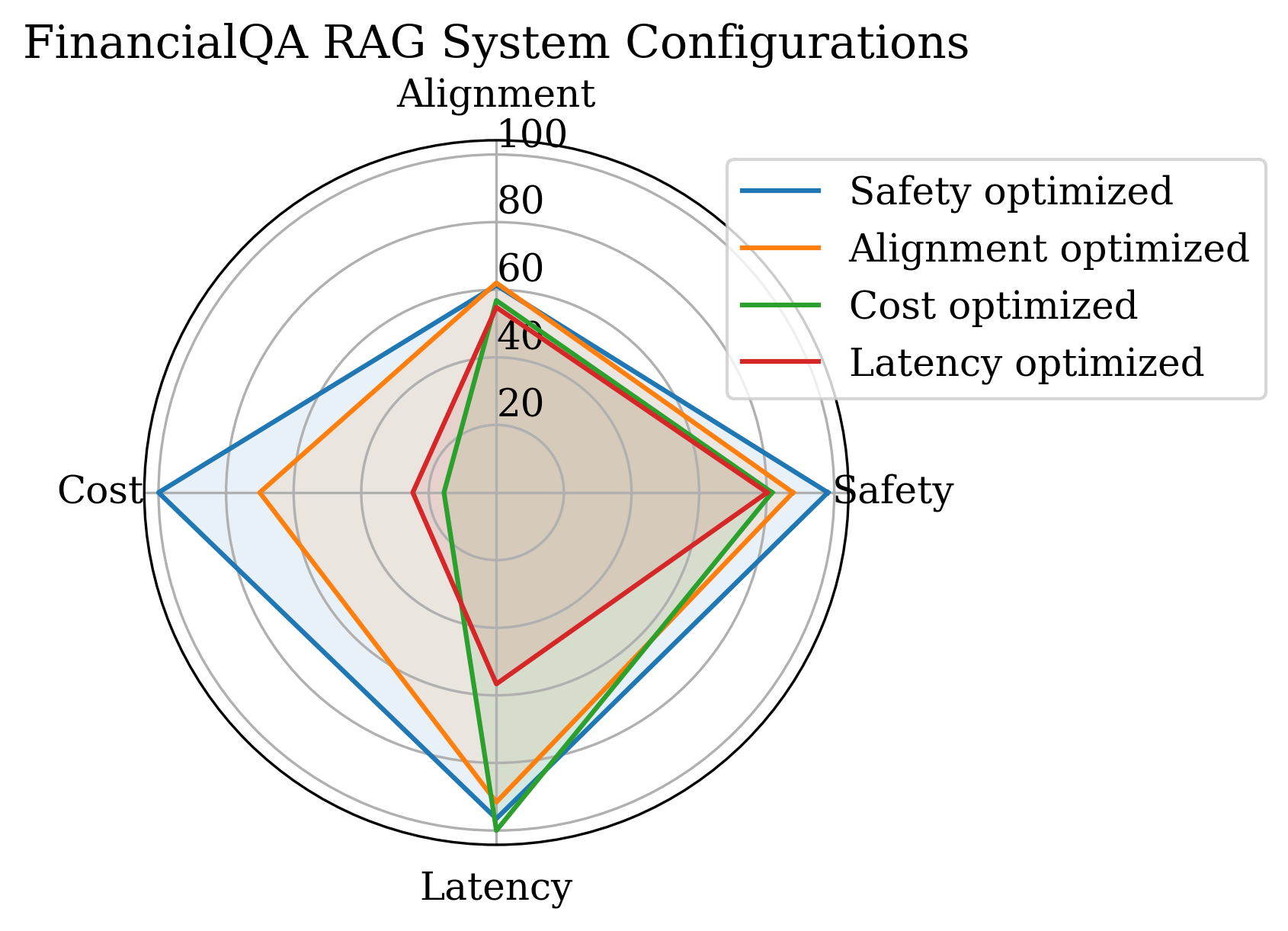}
   \includegraphics[width=0.4\linewidth]{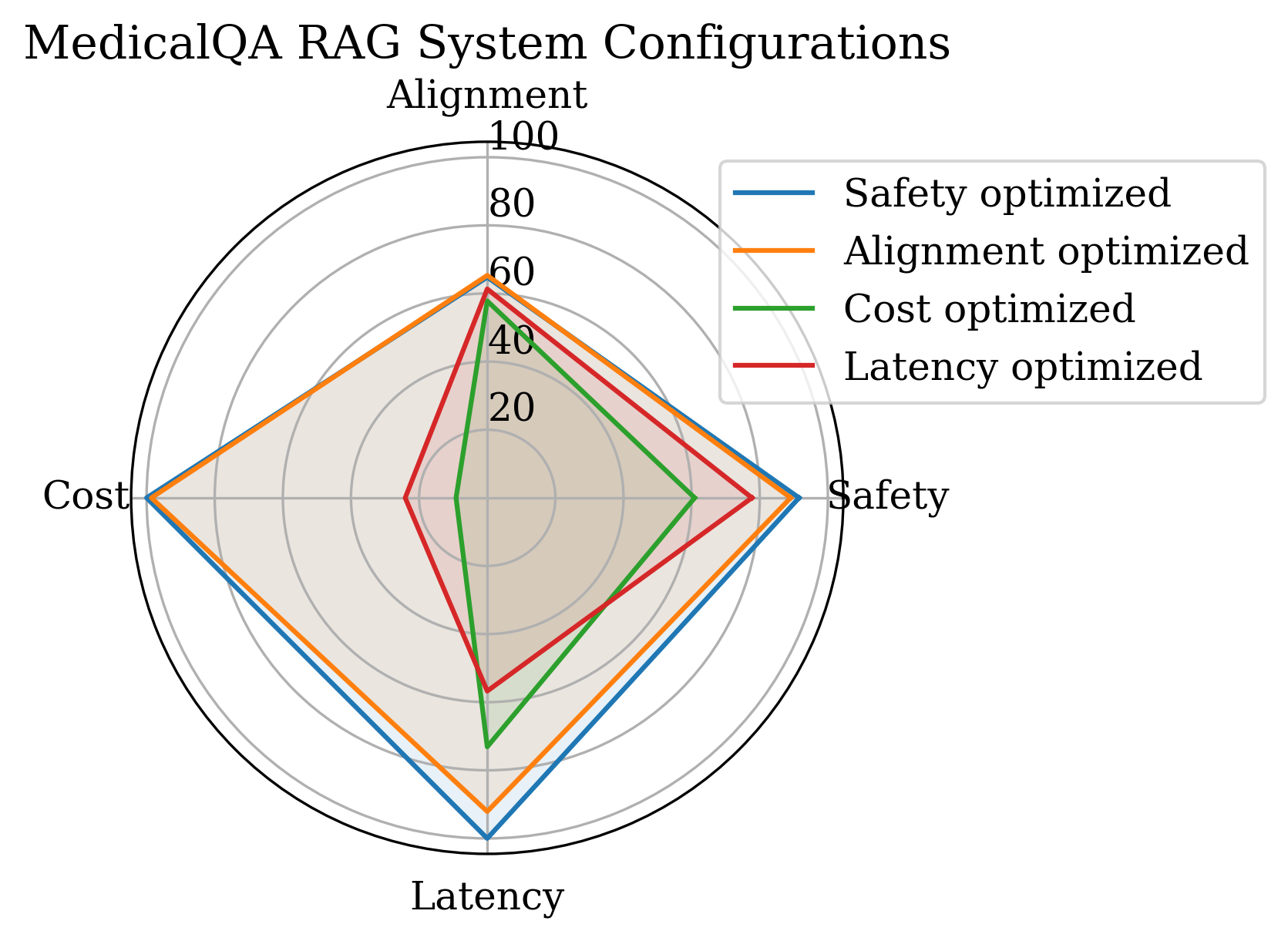}
    \caption{Radar charts comparing the four objective function evaluations for iterations chosen to optimize each objective. We see that improved safety can be achieved at the expense of increased cost and latency. \textbf{N.B. Lower is better for cost and latency but higher is better for safety and alignment.}}
    \label{fig:radar}
\end{figure}

We report the HV improvement on the train and test splits of both datasets, across five random seeds in Figure \ref{fig:hvi}. We find that BO methods significantly outperform other baseline approaches on both tasks, and that \texttt{qLogNEHVI} outperforms its noiseless variant. 
Both BO methods show significant improvement compared to baselines after iteration 20, when the BO acquisition function is used to select inputs rather than Sobol sampling. 

Figure \ref{fig:pareto-front} shows the Pareto front for a fixed seed for the FinancialQA and MedicalQA datasets. Since the overall Pareto frontier lies in $\mathbb{R}^4$ (as there are 4 objectives), we project onto three $\mathbb{R}^2$ plots for visualization purposes. We find that \texttt{qLogNEHVI} BO obtains a superior Pareto front compared to the baselines, also finding a wider spread of solutions. In particular, we notice significant clustering around sub-optimal solutions when using \texttt{qLogEHVI} BO, as observed by \cite{daulton2021parallel}.

Figure \ref{fig:radar} depicts the objective function evaluations across different configurations optimized for each objective. We see that safety and alignment optimized configurations come at the expense of cost and latency, and vice versa. We find that significant cost and latency reduction can be achieved at the cost of minimal safety and alignment reduction. All configuration settings and objective evaluations are detailed in Table \ref{tab:configurations}, where we observe similar chosen parameters across both datasets, especially the choice of LLM and embedding model. From these examples and others, we observe general patterns that help to optimize each objective. Safety optimized configurations often use the large embedding model and a large chunk size. In contrast, latency optimized configurations use the small embedding model and a small chunk size. This is intuitive: high safety requires sufficient high-quality context tokens, whereas low latency necessitates fewer context tokens. 
\section{Discussion}
\label{sec:discussion}
We frame our discussion as takeaways for industry practitioners that aim to optimize configurations of a RAG pipeline in a multi-objective setting. The first consideration
is what we call \emph{objective relationships}. In our experiments, we found that safety and alignment are often positively correlated with each other, and similarly for cost and latency. However, these two sets of objectives involve conflicting parameters, which makes it challenging to set a suitable trade-off between reliability (safety and alignment) and efficiency (cost and latency). Resolving conflicts between objectives is inherently challenging and remains an open question; we recommend that practitioners be thoughtful about latent relationships between objectives when choosing which objectives to optimize over. 

The second consideration is \emph{task dependence}, meaning that an optimal configuration for a RAG pipeline on a task in one setting may not generalize to another. While we observe configurations that work well across both tasks with respect to certain objectives (\eg high chunk size correlates with higher safety and alignment
), there is no optimal configuration that is shared across the two tasks. Practitioners should be aware that the optimal configuration will be highly dependent on the task they are building for, including the way their system will be used, the domain, and the stakeholders. 

The third consideration is \emph{objective dependence}, where objective evaluations follow different trends (or have no trend) across different configurations. For example, we see high objective dependence for temperature, since high temperatures (\eg $>1.0$) consistently reduced all four objectives. However, it is harder to discern a relationship between chunk overlap and the objectives, indicating low objective dependence. Task and objective dependence can also compound, highlighting the challenge of collectively optimizing the configuration of a RAG system.

\paragraph{Future Work} Our results demonstrate that end-to-end optimization of RAG systems is a promising avenue for research. We highlight two key areas for future exploration: first, improving the efficiency of Algorithm~\ref{alg:1}. A potential direction is ``decoupled evaluation'', where not all objectives are assessed at every iteration, towards a cost-aware optimization strategy. For instance, evaluating safety and alignment is significantly more expensive than measuring cost and latency.

Second, improvements can be made to the framework itself. Our current approach does not account for prompt engineering, despite its significant impact on response quality. Additionally, our safety and alignment metrics considered in this work remain limited in scope, excluding aspects such as toxicity and data leakage. However, our framework is inherently flexible and can be extended with adapted objectives or additional parameters. Another open challenge is configuration selection from the Pareto frontier. As Figure \ref{fig:pareto-front} illustrates, multiple configurations are Pareto-optimal, making the selection of the most suitable trade-off for a given application non-trivial.


\subsubsection*{Acknowledgments}
The authors thank Abhinav Raghunathan, Paul Dongha and David More for their guidance and industry expertise. We also thank Eirini Ioannou for her advice and suggestions. This work was supported in part by NSF awards 1916505, 1922658, 2312930, 2326193, and by the NSF GRFP (DGE-2234660).
This work was supported in part by ELSA: European Lighthouse on Secure and Safe AI project
(grant agreement No. 101070617 under UK guarantee). 
Views and opinions expressed are those of the author(s) only and do not necessarily reflect those of any of these funding agencies, European Union or European Commission.

\bibliography{main}
\bibliographystyle{iclr2025_conference}

\appendix
\section{Related Work on Fine-tuning and Model Merging}
\label{appendix:related}

While we do not explore fine-tuning or model-merging in our work, several authors have studied the effectiveness of HO to improve the alignment process. 

\paragraph{Fine-tuning.}\citet{wu2024beta} explored the importance of hyperparameter tuning the $\beta$-parameter\footnote{$\beta$ governs the extent to which the policy model's behavior can diverge from the original model.} for Direct Preference Optimization (DPO)~\citep{rafailov2024direct}, a popular framework for human-preference tuning LLMs. Significantly, they uncovered settings in which increasing $\beta$ improves DPO performance, and others where increasing $\beta$ has the exact opposite effect and decreases performance. 
\citet{wang2024llm} introduce an approach called Hyperparameter Aware Generation
(HAG), that allows LLMs to ``self-regulate'' hyperparameters like temperature, top-$p$, top-$k$, and repetition penalty during inference. They observed that different configurations of these hyperparameters lead to different performances on tasks like reasoning, creativity, translation, and math. 
\citet{tribes2023hyperparameter} used hyperparameter optimization (HO) to improve the instruction fine-tuning process, adjusting the hyperparameters rank and scaling $\alpha$ for Low-Rank Adaptation (LoRA)~\citep{hu2021lora}\footnote{Low-Rank Adaptation (LoRA) is a method that reduces the number of trainable parameters for a fixed model for downstream tasks like fine-tuning.}, as well as the model dropout rate and learning rate. In their experiments, they fine-tuned a Llama 2 7B parameter model\footnote{\url{https://huggingface.co/meta-llama/Llama-2-7b}}, and found that HO-fine-tuning resulted in better performance on tasks like MMLU~\citep{hendrycks2020measuring}, BBH~\citep{suzgun2022challenging}, DROP~\citep{dua2019drop}, and HumanEval~\citep{chen2021evaluating}, as compared to vanilla fine-tuning. Methodologically, they tested two HO approaches: Tree-structured Parzen Estimator tuning\footnote{TPE is a Bayesian optimization~\citep{bergstra2011algorithms} algorithm that uses a probabilistic model for HO. It is a Sequential Model-Based Optimization (SMBO) method~\citep{hutter2010sequential}.}~\citep{bergstra2011algorithms} and Mesh Adaptive Direct Search~\citep{audet2006mesh}, and found better performance with the latter. Overall, their results confirm the necessity of careful HO in instruction-tuning.

\paragraph{Model merging.} \citet{li2024s} frame LLM model merging, or combining different ``source'' (or base) models to create a unified model that retains the strengths of each model, as a multi-objective optimization (MOO) problem. They use parallel multi-objective Bayesian optimization (qEHVI)~\citep{daulton2020differentiable} to search over a range of model merging techniques like Model Soup~\citep{wortsman2022model} and TIES-Merging~\citep{yadav2024ties} (and the associated hyperparameters of those techniques), and evaluate the performance of the merged model on benchmarks like MMLU and Big-Bench Hard~\citep{suzgun2022challenging}. Our work is distinct from approaches using model merging in that we search over choices for LLMs rather than combine them. However, similar to ~\citet{li2024s}, we test the effectiveness of the qEHVI and qNEHVI~\citep{daulton2021parallel} (a variation allowing for noisy objectives) for HO.
\section{Approximating Objective Functions}
\label{sec:approx_objective}
Equation \ref{eq:objective} defines our proposed task as the maximization of the $\mathcal{HV}$ across multiple objectives. Each objective function evaluates a property of the RAG system for a given workload. We define a workload as a probability distribution across all possible queries, $P(q)$, where $q \in \mathcal{Q}$ is a user query for the RAG system. Further, we use $f_m^q: \mathcal{X}, \mathcal{Q} \mapsto \mathbb{R}$  to denote an evaluation function for an individual query and objective $m$. Using these definitions, we can write down the objective function as the expectation across queries:
\begin{equation}
    f_m(\mathbf{x}) = \sum_{q \in \mathcal{Q}} f_m^q(\mathbf{x}, q) P(q)
\end{equation}

Assuming we can sample queries from the workload, $q \sim P(q)$, we can use a Monte Carlo approximation for each objective function using a sample set $\mathcal{Q'} \subset \mathcal{Q}$:
\begin{equation}
\label{eq:mc}
    f_m(\mathbf{x}) \approx \frac{1}{|\mathcal{Q'}|}\sum_{q \in \mathcal{Q'}} f_m^q(\mathbf{x}, q)
\end{equation}
We assume that generating data synthetically using an LLM (Section \ref{sec:synthetic_data}) is equivalent to sampling queries $q \sim P(q)$ to obtain $\mathcal{Q'}$.
Our algorithm uses Equation \ref{eq:mc} as a tractable approximation of the objective functions in Equation \ref{eq:moo}.
\section{Experimental details}
\label{appendix:experiments}

We use the following search space for hyperparameters:
\begin{itemize}
    \item $c_s \in \mathbb{Z}^+$: Maximum number of tokens in each document chunk.
    \item $c_n \in \mathbb{Z}^+$: Number of chunks retrieved from the vector database for each query.
    \item $o \in \mathbb{Z}^+$: Number of tokens which overlap between adjacent chunks in a document.
    \item $t \in [0,1.2]$: Temperature of the LLM when generating responses.
    \item $r \in [0, 1]$: Rerank threshold used to set the minimum similarity between the context chunk and query, as evaluated by the reranker\footnote{We use a fixed rerank model \texttt{Salesforce/Llama-Rank-V1} provided by TogetherAI for all RAG systems.}. Retrieved documents which are below this threshold are ignored and not passed to the LLM as context. If no chunks exceed this threshold, we choose only the highest scoring chunk as context.
    \item $\ell \in \{\text{gpt-4o}, \text{gpt-4o-mini}, \text{llama-3.2-3B}, \text{llama-3.1-8B}\}$: Choice of LLM used to generate the response.
    \item $e \in \{\text{text-embedding-3-large},\text{text-embedding-3-small}
    \}$: Choice of embedding model when embedding the queries and document chunks.
\end{itemize}
\clearpage
\section{LLM Prompts}
\label{sec:prompts}
We use the following prompt with GPT-4o to generate synthetic questions from the financial source document:
\begin{lstlisting}
You are an expert synthetic data generation agent. Look at this PDF document containing information on accounting from a leading global financial services organization. Generate 50 questions which accountants might ask based only on the information provided in the document. Example questions are:

Are there any specific disclosures required when transferring HTM securities to AFS?
Are there any tax implications to consider when transferring securities between categories?
What are the steps to recording a transfer of AFS securities to HTM?
How should Federal agencies implement the new Land standard? 
I have a client in bankruptcy, what is the presentation of financial statements once they emerge from bankruptcy
How do you determine if a limited partnership is a VIE
can a debt being refinanced with a different lender result in modification accounting?
Are there specific criteria for capitalizing costs related to PP&E additions?
how do you account for PP&E additions
which examples in the debt and equity handbook illustrate the accounting for preferred stock?
Are there any examples of exit fees in investment company accounting?
\end{lstlisting}

For prompting the LLM during RAG, we use the following prompt from \url{https://smith.langchain.com/hub/rlm/rag-prompt}:

\begin{lstlisting}
You are an assistant for question-answering tasks. Use the following pieces of retrieved context to answer the question. If you don't know the answer, just say that you don't know. Use three sentences maximum and keep the answer concise.
Question: {question} 
Context: {context} 
Answer:
\end{lstlisting}
\clearpage
\section{Additional Results}
\begin{table}[h]
    \centering
    \begin{tabular}{l l p{1.8cm} p{1.8cm} p{1.8cm} p{1.8cm}}
    \toprule
    & & \multicolumn{4}{c}{\textbf{FinancialQA Optimized Configurations}} \\ 
    & & \textbf{Safety} & \textbf{Alignment} & \textbf{Cost} & \textbf{Latency} \\ \midrule
    \multirow{7}{*}{Hyperparam}& Embedding model & text-embedding-3-large & text-embedding-3-large & text-embedding-3-large & text-embedding-3-small\\
    & LLM & gpt-4o-mini & Llama-3.1-8B & Llama-3.2-3B & Llama-3.1-8B\\
    & Chunk size & 1024 & 1024 & 512 & 1024\\
    & Chunk overlap & 512 & 128 & 64 & 64 \\
    & Num chunks & 3 & 4 & 3 & 3 \\
    & Rerank threshold & 0.00 & 0.64 & 0.24 & 1.00 \\
    & Temperature & 0.03 & 0.12 & 0.79 & 0.00 \\ \midrule
    \multirow{4}{*}{Objective} & Safety & \textbf{98.1} & \underline{87.8} & 81.6 & 80.1 \\
    & Alignment &  \underline{61.3} & \textbf{62.0} & 56.8 & 54.8 \\
    & Cost & 585 & 410 & \textbf{90.8} & \underline{145} \\
    & Latency & 12.4 & \underline{11.7} & 12.8 & \textbf{7.26} \\
    \bottomrule
    \end{tabular}

    \begin{tabular}{l l p{1.8cm} p{1.8cm} p{1.8cm} p{1.8cm}}
    \toprule
    & & \multicolumn{4}{c}{\textbf{MedicalQA Optimized Configurations}} \\ 
    & & \textbf{Safety} & \textbf{Alignment} & \textbf{Cost} & \textbf{Latency} \\ \midrule
    \multirow{7}{*}{Hyperparam}& Embedding model & text-embedding-3-large & text-embedding-3-small & text-embedding-3-large & text-embedding-3-small\\
    & LLM & gpt-4o-mini & gpt-4o-mini & Llama-3.1-8B & Llama-3.1-8B\\
    & Chunk size & 1024 & 1024 & 256 & 1024\\
    & Chunk overlap & 256 & 512 & 32 & 32 \\
    & Num chunks & 6 & 6 & 2 & 2 \\
    & Rerank threshold & 0.00 & 0.22 & 0.57 & 0.36 \\
    & Temperature & 0.00 & 0.10 & 0.57 & 0.00 \\ \midrule
    \multirow{4}{*}{Objective} & Safety & \textbf{91.5} & \underline{89.1} & 60.9 & 77.7 \\
    & Alignment &  \underline{64.8} & \textbf{65.3} & 57.8 & 61.3 \\
    & Cost & 1010 & 997 & \textbf{92.6} & \underline{244} \\
    & Latency & 17.0 & 15.6 & \underline{12.4} & \textbf{9.62} \\
    \bottomrule
    \end{tabular}
    \caption{Input parameters and objective evaluations for individual configurations optimized for each objective. We observe a similar choice of parameters between both datasets, especially the choice of LLM and embedding model.}
    \label{tab:configurations}
\end{table}

\end{document}